\documentclass[runningheads]{llncs}
\usepackage[T1]{fontenc}
\usepackage[utf8]{inputenc} 

\usepackage{graphicx}
\usepackage{pifont} 
\usepackage{amsmath} 
\usepackage{subcaption}
\usepackage{amssymb} 

\newcommand{\cmark}{\ding{51}}

\usepackage{color}
\usepackage{hyperref} 

\begin{document}
\title{Lightweight Prompt-Guided CLIP Adaptation for Monocular Depth Estimation}
\titlerunning{Lightweight Prompt-Guided CLIP Adaptation}
%
\author{Reyhaneh Ahani Manghotay\inst{1} \and
Jie Liang\inst{2}\thanks{Corresponding author.}}
\authorrunning{R. Ahani Manghotay and J. Liang}
\institute{School of Engineering Science, Simon Fraser University, Burnaby, BC, Canada\\
\email{raa112@sfu.ca} \and
School of Information Science and Technology, Eastern Institute of Technology, Ningbo, China\\
\email{Jliang@eitech.edu.cn}}

\maketitle              
\begin{abstract}
Leveraging the rich semantic features of vision-language models (VLMs) like CLIP for monocular depth estimation tasks is a promising direction, yet often requires extensive fine-tuning or lacks geometric precision. We present a parameter-efficient framework, named \textbf{MoA-DepthCLIP}, that adapts pretrained CLIP representations for monocular depth estimation with minimal supervision. Our method integrates a lightweight Mixture-of-Adapters (MoA) module into the pretrained Vision Transformer (ViT-B/32) backbone combined with selective fine-tuning of the final layers. This design enables spatially-aware adaptation, guided by a global semantic context vector and a hybrid prediction architecture that synergizes depth bin classification with direct regression. To enhance structural accuracy, we employ a composite loss function that enforces geometric constraints. On the NYU Depth V2 benchmark, MoA-DepthCLIP achieves competitive results, significantly outperforming the DepthCLIP baseline by improving the $\delta_1$ accuracy from 0.390 to 0.745 and reducing the RMSE from 1.176 to 0.520. These results are achieved while requiring substantially few trainable parameters, demonstrating that lightweight, prompt-guided MoA is a highly effective strategy for transferring VLM knowledge to fine-grained monocular depth estimation tasks.

\end{abstract}
\section{Introduction}

The advent of VLMs, most notably CLIP \cite{radford2021learning}, has revolutionized the fields of computer vision and natural language processing. By learning joint representations from vast quantities of image-text pairs, these models have demonstrated remarkable capabilities in zero-shot and few-shot learning for tasks like image classification \cite{gao2021clip,zhang2021tip,zhou2021learning} and have been successfully adapted for dense prediction problems such as object detection and segmentation \cite{rao2021denseclip,zhou2022detic}. Their influence even extends to 3D perception, with applications in point cloud understanding \cite{zhang2022pointclip}. However, a fundamental challenge remains: translating the high-level, semantic knowledge encapsulated in VLMs into the fine-grained, metric predictions required for geometric tasks like monocular depth estimation.

Monocular depth estimation is a fundamental task in computer vision, providing critical information for applications ranging from autonomous navigation and robotics to augmented reality. It is also a foundational step for other complex tasks, including monocular 3D object detection and single-image point cloud reconstruction \cite{zeng2018pc,zhang2022monodetr}. Historically, progress in this domain has been driven by fully supervised methods \cite{fu2018deep,li2022depthformer} that, while accurate, depend on large-scale datasets with dense depth annotations (e.g., NYU Depth V2 \cite{silberman2012indoor}), which are known to be expensive and time-consuming to create. To mitigate this data dependency, a new paradigm has emerged: large-scale, foundation-style models \cite{bhat2023zoedepth,li2023metric3d,baek2024unidepth,yang2024depthanything}. These models achieve state-of-the-art performance by training on diverse, aggregated datasets but introduce substantial computational and parameter burdens, limiting their practical deployment. This leaves a critical research gap for methods that are both data- and compute-efficient.

An alternative direction, first explored by DepthCLIP \cite{zhang2022depthclip}, leveraged VLM alignment for zero-shot depth estimation. By reformulating depth estimation as a language-driven classification problem, matching image regions to textual prompts like "close" or "far", DepthCLIP provided a proof-of-concept without any task-specific training. While it is novel, its reliance on handcrafted prompts and coarse depth discretization limited its practical utility, producing outputs that lacked geometric detail. Subsequent efforts sought to overcome these limitations, focusing on strategies like learning more effective prompts \cite{auty2023learning,kim2024clip}, creating adaptive, image-aware depth bins \cite{son2024cabins}, and developing efficient few-shot adaptation mechanisms \cite{hu2024learning}.

Motivated by these challenges, we propose \textbf{MoA-DepthCLIP}, a framework that bridges the semantic power of CLIP with the geometric precision needed for depth estimation, while maintaining exceptional efficiency. Our core contribution is a novel synthesis of two powerful, yet previously separate, paradigms.

First, while MoA has proven effective for parameter-efficient tuning in other domains like NLP \cite{wang2022adamix,liu2025dm-adapter}, its application to dense geometric tasks like monocular depth estimation remains unexplored. Second, while hybrid classification-regression architectures are a known standard in traditional depth estimation \cite{fu2018deep,bhat2021adabins}, they have not been integrated with modern, lightweight VLM adaptation strategies like MoA.

Our approach \textbf{integrates} these two concepts: we introduce a lightweight MoA module and selective fine-tuning into the ViT backbone, and feed the resulting adapted features into this proven hybrid head architecture. This novel combination enables parameter-efficient, spatially-aware adaptation to the depth task. Our model is guided by a global scene context vector, derived from averaged text prompt embeddings, and enforced by a composite loss function that combines classification and regression losses (L1 and scale-invariant terms).

Our contributions are summarized as follows:
\begin{itemize}
\item We introduce \textbf{MoA-DepthCLIP}, the first adaptation strategy for monocular depth estimation based on a lightweight MoA PEFT approach, combined with selective backbone fine-tuning.
    \item We demonstrate how to \textbf{integrate} this modern, VLM-native adaptation strategy with a classic, geometry-focused hybrid (classification-regression) prediction head to recover fine-grained metric details.
    \item We demonstrate through experiments on NYU Depth V2 that our synthesized approach significantly outperforms prior VLM-based approaches like DepthCLIP, improving $\delta_1$ accuracy from 0.390 to 0.745 and reducing RMSE by over 55\%, while using only a fraction of the trainable parameters typically required by foundation models.
\end{itemize}

\begin{figure}[t]  
\centering
\includegraphics[width=\textwidth]{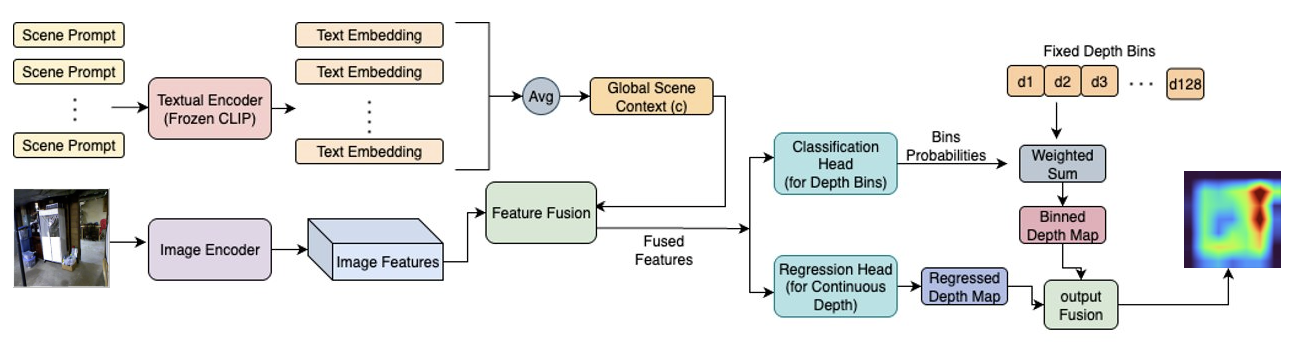}
\caption{\textbf{Overall architecture of MoA-DepthCLIP.} Scene prompts are encoded using a frozen CLIP text encoder to form a global scene context vector. The image is encoded with a frozen ViT-B/32 backbone augmented with MoAs. Fused features are then passed to a dual-head prediction module: one head performs depth bin classification and produces a binned depth map via weighted summation, while the other head performs direct regression. The final output depth map is a fusion of both predictions.}
\label{fig:pipeline}
\end{figure}

\section{Related Work}

\subsection{Vision-Language Models}
The development of large-scale vision-language models, particularly CLIP \cite{radford2021learning}, marked a significant milestone by learning powerful, transferable representations from image-text pairs. The success of CLIP in zero-shot classification \cite{gao2021clip,zhang2021tip,zhou2021learning} inspired adaptations for dense prediction tasks. For instance, DenseCLIP \cite{rao2021denseclip} demonstrated how to effectively use CLIP's features for semantic segmentation through careful prompt engineering. While these models show a strong grasp of semantic concepts, applying this knowledge to fine-grained geometric tasks like depth estimation remains a key challenge.

\subsection{Monocular Depth Estimation}
Depth estimation from a single image is a long-standing task in computer vision. Supervised methods \cite{fu2018deep,li2022depthformer} have consistently pushed the boundaries of accuracy, but their reliance on large-scale, annotated datasets like NYU Depth V2 \cite{silberman2012indoor} is a significant bottleneck. In parallel, self-supervised approaches have sought to alleviate this by using geometric constraints from video sequences as a supervisory signal \cite{mahjourian2018unsupervised,zhang2020unsupervised}. More recently, the trend has shifted towards "foundation models" that achieve state-of-the-art generalization by training on massive, diverse datasets, though this performance comes at a high computational cost \cite{bhat2023zoedepth,li2023metric3d,baek2024unidepth,yang2024depthanything}.

\subsection{Vision-Language for Depth Estimation}
Directly leveraging vision-language models for depth estimation is a relatively new direction. The first major work in this area is DepthCLIP \cite{zhang2022depthclip}, which introduced a zero-shot approach by reframing depth prediction as a classification problem. It matches image regions to handcrafted text prompts like "close" or "far" to generate a coarse depth map without any fine-tuning. While this was a novel proof-of-concept, its practical use was limited by the coarse nature of its output and its dependence on manually designed prompts. 

Subsequent efforts sought to overcome these limitations, focusing on strategies like learning more effective prompts \cite{auty2023learning,kim2024clip}, creating adaptive, image-aware depth bins \cite{son2024cabins}, and developing efficient few-shot adaptation mechanisms \cite{hu2024learning}.

Our work also builds upon this idea, but significantly advances the paradigm beyond simple zero-shot prompting. We introduce \textbf{MoA-DepthCLIP}, a framework that replaces static prompt engineering with a learnable, lightweight adaptation strategy using MoA. Furthermore, we address the geometric limitations of prior work by integrating a global scene context and a fine-grained hybrid prediction head. This holistic approach effectively bridges the gap between CLIP's high-level semantic understanding and the precise, metric requirements of monocular depth estimation.



\section{Method}
\subsection{Mixture of Adapters}

MoA-DepthCLIP method introduces a parameter-efficient strategy to adapt the pretrained CLIP \cite{radford2021learning} visual backbone, which is based on the ViT-B/32 architecture \cite{dosovitskiy2020vit}, by integrating Mixture-of-Adapters (MoA) modules into its transformer layers. An overview of our full architecture, which illustrates the placement of these modules, is depicted in Figure~\ref{fig:pipeline}. Each MoA module consists of three main components: a set of lightweight experts, a gating network that determines token-specific routing weights, and a residual mixing operation that injects the adapted features back into the backbone.

\begin{figure}[t]
\centering

\begin{subfigure}[b]{0.55\textwidth} 
    \centering
    \includegraphics[width=0.90\linewidth]{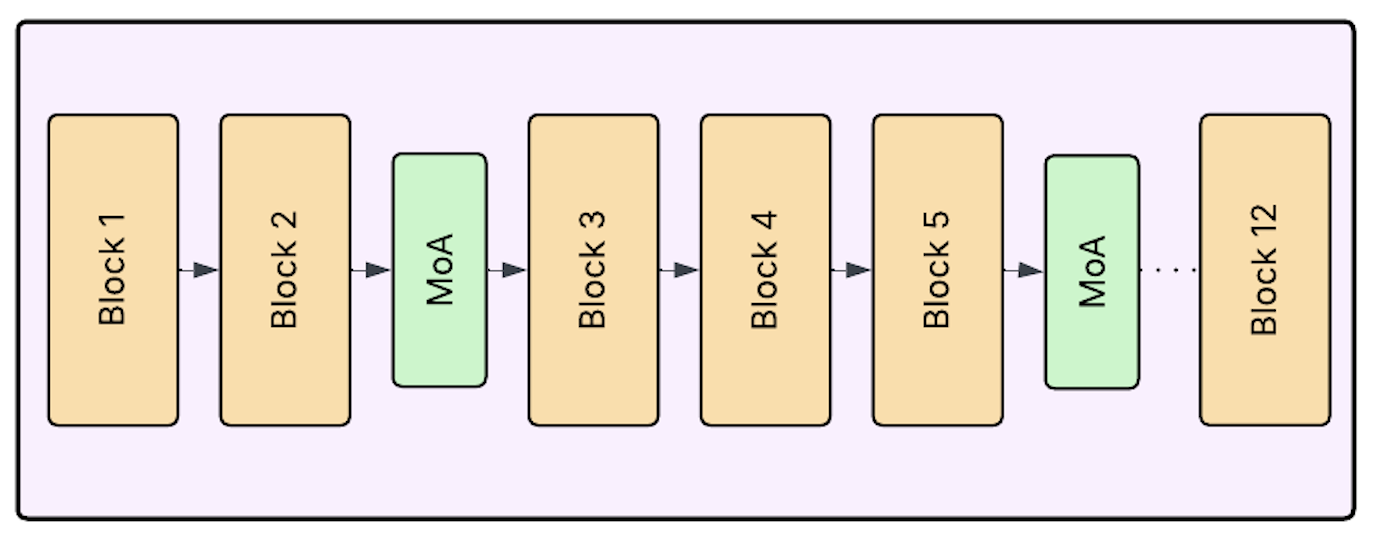} 
    \caption{Selective MoA Placement. We insert MoA modules (green) at key layers ({2, 5, 8, 11}) of the ViT-B/32 encoder (tan).}
    \label{fig:vit_placement}
\end{subfigure}
\hfill 
\begin{subfigure}[b]{0.4\textwidth}
    \centering
    \includegraphics[width=\textwidth]{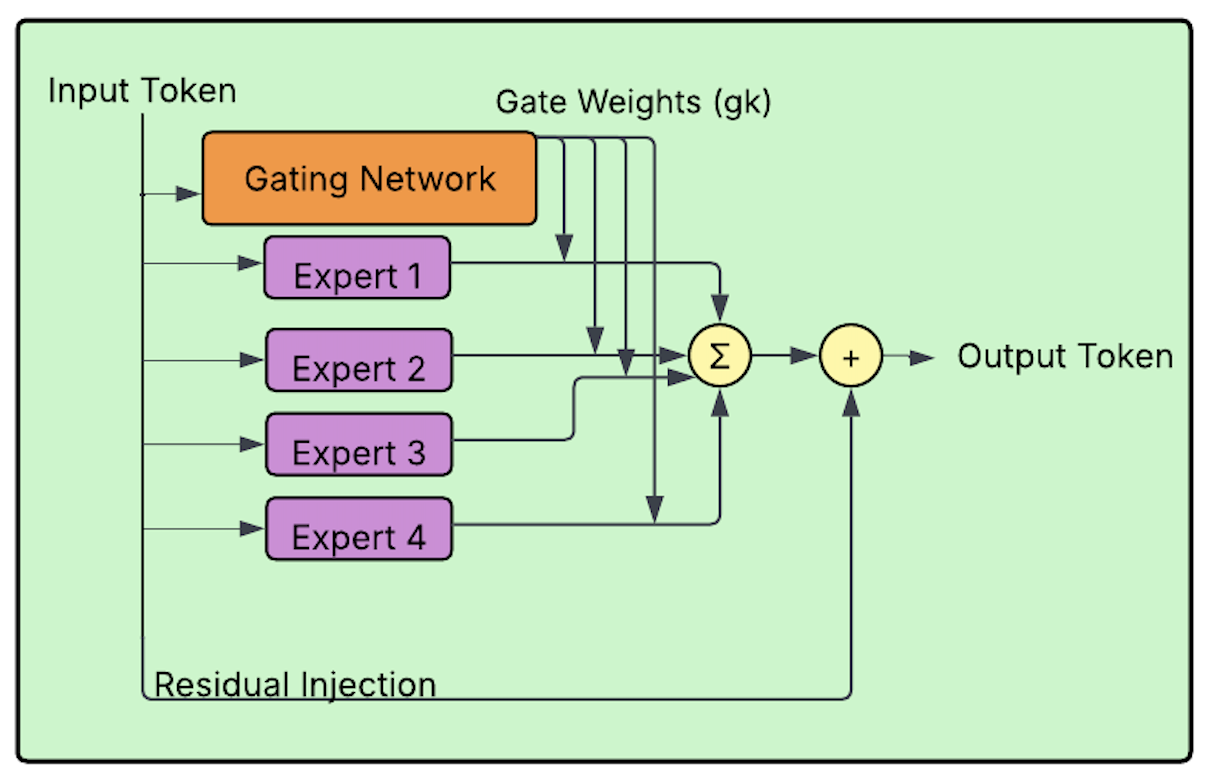}
    \caption{Internal architecture of a single MoA module, showing the Gating Network, K=4 Experts, and residual injection.}
    \label{fig:moa_detail}
\end{subfigure}

\caption{Overview of our lightweight adaptation strategy. (a) Illustrates the selective placement of MoA modules within the ViT backbone. (b) Details the internal architecture of a single MoA module.}
\label{fig:method_overview}
\end{figure}

\paragraph{Expert Architecture.}  
Each expert is implemented as a two-layer MLP with a bottleneck structure:
\begin{equation}
\text{Expert}(x) = W_2 \, \sigma(W_1 x),
\label{eq:expert}
\end{equation}
where \( W_1 \in \mathbb{R}^{d \times d_b} \), \( W_2 \in \mathbb{R}^{d_b \times d} \), \( d \) is the transformer hidden dimension (768 for ViT-B/32), \( d_b = 64 \) is the adapter bottleneck size, and \(\sigma\) is the GELU activation. This design follows the standard adapter formulation but is kept deliberately small to minimize parameter overhead. 

\paragraph{Gating Network.}
To enable token-specific specialization, we employ a gating network, similar in principle to those used in Mixture-of-Experts PEFT approaches like AdaMix \cite{wang2022adamix}. For each token representation \( x_i \), the network predicts routing probabilities over the \( K \) experts using a standard softmax function:
\begin{equation}
g_i = \text{softmax}\left(\frac{G(x_i)}{\tau}\right),
\label{eq:gating}
\end{equation}
where \( G(\cdot) \) is a two-layer MLP (768 → 128 → \(K\)) with ReLU activation, and \(\tau = 2.0\) is a temperature hyperparameter. Our key difference from AdaMix \cite{wang2022adamix}, which uses stochastic routing during training and expert merging for inference, lies in our direct and deterministic use of these probabilities. We use the computed \( g_i \) values directly to form a weighted combination of expert outputs (Eq.~\ref{eq:mixture}) during both training and inference. This constitutes our contribution for this component: applying a simpler, deterministic gating mechanism within the MoA framework for stable adaptation in depth estimation.

\paragraph{Mixture and Residual Injection.}
Following our deterministic gating mechanism, the outputs of all \(K\) experts for a given token \(x_i\) are combined via a weighted sum using the gate probabilities \(g_{i,k}\) from Eq.~\ref{eq:gating}. Importantly, this mixture is then added back to the original token representation using a residual connection, a standard technique in adapter modules \cite{houlsby2019adapter} to ensure training stability and preserve the valuable features learned during pre-training:
\begin{equation}
\tilde{x}_i = x_i + \sum_{k=1}^K g_{i,k} \, \text{Expert}_k(x_i).
\label{eq:mixture}
\end{equation}
We leverage this residual design within MoA-DepthCLIP to seamlessly integrate specialized expert adaptations while strictly preserving the backbone's original representational power, a balance critical for fine-grained tasks like depth estimation. Furthermore, during training, we monitor expert usage derived from \(g_{i,k}\) to ensure effective load balancing, differentiating our approach from simpler adapter methods that lack this dynamic specialization.

\paragraph{Selective Layer Integration.}
A key design choice in our work is the selective placement of the MoA modules. Instead of inserting modules uniformly after every transformer block, we adopt a targeted strategy. We empirically determined that inserting MoA modules at only four key transformer blocks of the ViT-B/32 encoder, specifically layers $\{2, 5, 8, 11\}$, effectively balances adaptation across early, mid-level, and late features. This sparse placement strategy constitutes a specific contribution of our method, optimizing performance while keeping the parameter count minimal.

\paragraph{Expert Diversity.}
A potential challenge in MoE and MoA systems is expert collapse, where the gating network predominantly routes tokens to only a few experts, negating the benefits of specialization \cite{wang2022adamix}. While some methods introduce auxiliary loss terms during training to enforce load balancing, we adopt a simpler approach focused on monitoring. We track the entropy of the gating distribution (Eq.~\ref{eq:entropy}) during training:
\begin{equation}
H(g_i) = - \sum_{k=1}^K g_{i,k} \log g_{i,k}.
\label{eq:entropy}
\end{equation}
A high average entropy indicates that multiple experts are being actively utilized. Our contribution here is using entropy solely as a diagnostic tool to verify effective expert usage and training stability, rather than incorporating complex load-balancing losses. This monitoring confirms that our MoA implementation achieves the desired specialization without collapse.

Overall, our adaptation strategy, which combines lightweight MoA modules with selective fine-tuning of the final four backbone layers, enables spatially-aware, token-level specialization. While the MoA modules themselves add a negligible number of parameters, the total number of trainable parameters in our model remains a small fraction of the full backbone, making our approach a highly efficient alternative to full fine-tuning.

\subsection{Global Scene Context Fusion}
The original DepthCLIP \cite{zhang2022depthclip} framework relied on simple, handcrafted prompts applied at the pixel level (e.g., ‘close’, ‘far’) to represent coarse depth categories. While innovative for zero-shot prediction, this approach lacks sensitivity to the broader scene context. To provide a stronger, yet simpler, semantic prior compared to methods that learn continuous prompts \cite{zhou2021learning}, we introduce a mechanism to fuse a fixed, global scene context vector with the visual features. MoA-DepthCLIP leverages the CLIP text encoder \cite{radford2021learning} in a parameter-free manner. To provide a semantic prior relevant to our target benchmark (NYU Depth V2 \cite{silberman2012indoor}), we define a fixed set of text prompts aligned with its common indoor scene categories (e.g., “a photo of a kitchen,” “a photo of a classroom,” etc.). These prompts are encoded using the frozen CLIP text encoder, and their embeddings are $\ell_2$-normalized. We compute a single, generic context vector, $\mathbf{c}$, representing a general notion of "indoor scenes," by element-wise averaging these prompt embeddings. This strategy establishes a constant semantic anchor throughout training and inference without introducing any learnable parameters, ensuring consistent guidance for the visual features.

This global context vector $\mathbf{c}$ is then spatially fused with the visual feature map from the MoA-adapted backbone. We broadcast $\mathbf{c}$ to match the spatial dimensions of the visual features and concatenate them along the channel dimension. This global fusion differs significantly from DepthCLIP's pixel-wise prompt matching and prompt learning's adaptable vectors, providing instead a uniform, high-level prior across the entire image that complements the local visual details. The resulting fused representation serves as input to the subsequent hybrid prediction module.

\subsection{Depth Binning Strategy}
We adopt a hybrid prediction head involving depth bin classification, building on the effectiveness of depth discretization techniques \cite{fu2018deep}. Specifically, we predict a per-pixel distribution over $N$ discrete depth bins, following the general formulation also used by \cite{zhang2022depthclip}.
However, we significantly differ from DepthCLIP \cite{zhang2022depthclip} in our approach to defining these bins. DepthCLIP employed a small number (10) of fixed, handcrafted bins based on semantic distances. Recognizing that bin granularity is crucial, but aiming for a simpler approach than methods predicting per-image adaptive bins like AdaBins \cite{bhat2021adabins}, we systematically explored different counts for fixed bins. Through ablation studies (detailed in Section~\ref{sec:ablation}), we found that using $N=128$ bins provided the optimal trade-off between accuracy and robustness specifically for our MoA-adapted CLIP framework. We empirically demonstrate that adopting a fixed bin count of $N=128$ yields substantial improvement over the coarse discretization of DepthCLIP without requiring the computational complexity of adaptive binning. This configuration provides the optimal trade-off between fine-grained accuracy and model robustness in the context of VLM adaptation. We therefore adopt $N=128$ bins for all experiments.

\subsection{Composite Loss Function}
\label{sec:composite_loss}

To train our hybrid architecture with parallel classification and regression heads, we employ a composite loss function. This strategy of combining losses suited for classification (for stability) and regression (for detail) is a well-established practice in modern depth estimation frameworks \cite{fu2018deep,bhat2021adabins}. Our total loss is a weighted sum of three standard components:
\begin{equation}
\mathcal{L}_{\text{total}} = \lambda_{\text{cls}} \mathcal{L}_{\text{cls}} + \lambda_{\text{reg}} \mathcal{L}_{\text{reg}} + \lambda_{\text{silog}} \mathcal{L}_{\text{silog}},
\label{eq:total_loss}
\end{equation}
where each term targets a specific aspect of the prediction. We tailor this weighting configuration to stabilize the training of our lightweight MoA experts, ensuring they capture both high-level structural layout and low-level metric details.

\paragraph{Classification Loss ($\mathcal{L}_{\text{cls}}$).}
To supervise the classification head, we use a standard per-pixel Cross-Entropy (CE) loss. This loss encourages the model to predict the correct discrete depth bin for each pixel by comparing the output logits with the ground-truth bin indices. It provides a strong, stable supervisory signal that is particularly effective for learning the coarse layout of a scene.

\paragraph{Regression Losses.}
The regression head, which predicts a continuous depth map, is supervised by two complementary loss terms.

\textbf{L1 Loss ($\mathcal{L}_{\text{reg}}$):} We employ a per-pixel L1 loss, which penalizes the absolute difference between the predicted depth $\hat{d}_i$ and the ground-truth depth $d_i$. This provides a direct, fine-grained signal for improving geometric accuracy at a local level.

\textbf{Scale-Invariant Logarithmic Loss ($\mathcal{L}_{\text{silog}}$):} To ensure robustness against the global scale and shift ambiguities inherent in monocular depth estimation, we use the Scale-Invariant Logarithmic (SILog) loss, inspired by the formulation in \cite{eigen2014depth}. For each pixel $i$ in a valid mask, the loss is defined based on the variance and mean of the log-differences $g_i = \log \hat{d}_i - \log d_i$:
\begin{equation}
\mathcal{L}_{\text{silog}} = \alpha \left( \text{Var}(g) + \lambda \text{Mean}(g)^2 \right),
\label{eq:silog}
\end{equation}
where $\text{Var}(g)$ and $\text{Mean}(g)$ are computed over all valid pixels, $\lambda=0.85$ is a weighting term, and $\alpha=10.0$ is a scaling factor.

\paragraph{Loss Weighting.}
The components are combined using fixed weights to balance their contributions. Based on our experimental setup, we set the weights to $\lambda_{\text{cls}} = 1.0$, $\lambda_{\text{reg}} = 1.0$, and $\lambda_{\text{silog}} = 0.5$. This weighting scheme provides a balanced objective that prioritizes both coarse categorical accuracy from the classification head and fine-grained geometric fidelity from the regression head.

\section{Experiments}

\subsection{Dataset and Evaluation Metrics}
We evaluate our framework on the NYU Depth V2 dataset \cite{silberman2012indoor}, a widely used indoor RGB-D benchmark captured by a Kinect sensor. Depth values are capped at 10 meters following standard protocol. 

For evaluation, we report commonly used metrics including the Absolute Relative Error (AbsRel), Root Mean Squared Error (RMSE), log10 error, and threshold accuracies $\delta_1$, $\delta_2$, $\delta_3$. Here, $\delta_i$ measures the fraction of pixels with predictions within a factor of $1.25^i$ of the ground truth. Together, these metrics capture global accuracy and scale robustness.

\subsection{Implementation Details}
MoA-DepthCLIP is implemented in PyTorch, using the pretrained ViT-B/32 from OpenCLIP \cite{radford2021learning} as the visual backbone. To balance feature preservation and task adaptation, we keep most of the backbone frozen but fine-tune the final 4 Transformer blocks. The primary trainable components of our model are these unfrozen blocks, the MoA modules, and the two prediction heads (classification and regression). The global context vector, derived from text prompts, remains fixed during training.

We use $K=4$ experts for each MoA module with a bottleneck dimension of $d_b=64$. The model is trained for 30 epochs with a batch size of 8 on a single NVIDIA H100 GPU. We use the AdamW optimizer with a constant learning rate of $1 \times 10^{-5}$ and a weight decay of $1 \times 10^{-4}$. The number of depth bins for the classification head is set to $N=128$, which we found to yield the best performance in our ablation studies (see Section~\ref{sec:ablation}).

\subsection{Quantitative Results}
Table~\ref{tab:main_results} summarizes the results of our ablation study, showing the progressive impact of each component of our framework, starting from a reproduced DepthCLIP baseline.

\begin{table}[t]
\centering
\caption{\textbf{Quantitative results on NYU Depth V2.} We evaluate different configurations by progressively introducing composite loss, MoA, and increasing the number of bins. The final configuration, \textbf{MoA-DepthCLIP} (with ViT-B/32, MoA, composite loss, and 128 bins), achieves the best performance across all metrics.}
\label{tab:main_results}
\small
\renewcommand{\arraystretch}{1.15}
\setlength{\tabcolsep}{6pt}
\resizebox{\textwidth}{!}{
\begin{tabular}{l|c|c|c|c|ccc|ccc}
\hline
\textbf{Method} & \textbf{Backbone} & \textbf{Comp. Loss} & \textbf{MoA} & \textbf{\#Bins} 
& $\boldsymbol{\delta_1 \uparrow}$ & $\boldsymbol{\delta_2 \uparrow}$ & $\boldsymbol{\delta_3 \uparrow}$ 
& \textbf{AbsRel $\downarrow$} & \textbf{log10 $\downarrow$} & \textbf{RMSE $\downarrow$} \\
\hline
\textsc{DepthCLIP}       & ResNet-50 &  &  & 10  & 0.390  & 0.680  & 0.848  & 0.393  & 0.158  & 1.176 \\
\textsc{Our Baseline}              & ViT-B/32  &  &  & 10  & 0.417  & 0.701  & 0.890  & 0.377  & 0.147  & 1.096 \\
\textsc{+ Composite Loss}          & ViT-B/32  & \cmark &  & 10  & 0.503  & 0.791  & 0.925  & 0.310  & 0.121  & 0.843 \\
\textsc{+ MoA}                     & ViT-B/32  & \cmark & \cmark & 10  & 0.508  & 0.797  & \textbf{0.931}  & \textbf{0.308}  & 0.120  & 0.821 \\
\textbf{\textsc{MoA-DepthCLIP}} 
                                     & ViT-B/32  & \cmark & \cmark & 128 & \textbf{0.745} & \textbf{0.841} & {0.895} & {0.321} & \textbf{0.098} & \textbf{0.520} \\
\hline
\end{tabular}
}
\end{table}

Our analysis reveals a clear trajectory of improvement. First, simply replacing the original ResNet-50 backbone with ViT-B/32 provides a notable performance gain. The most significant leap occurs with the introduction of our composite loss function, which drastically improves accuracy (e.g., boosting $\delta_1$ from 0.417 to 0.503). Subsequently, integrating the MoA modules yields further incremental gains. Finally, optimizing the number of depth bins to 128 leads to another substantial improvement, particularly in the threshold accuracies and RMSE. MoA-DepthCLIP significantly outperforms the reproduced DepthCLIP baseline, improving the $\delta_1$ accuracy from 0.390 to 0.745 and reducing the RMSE from 1.176 to 0.520.

A noteworthy trade-off becomes apparent when analyzing this final step. While increasing the number of bins to 128 yields a dramatic improvement in the stricter accuracy thresholds ($\boldsymbol{\delta_1}$ and $\boldsymbol{\delta_2}$), we observe a slight decrease in the most lenient one ($\boldsymbol{\delta_3}$). We attribute this to the increased specialization of the final model. By operating in a much finer-grained prediction space, the model makes highly precise predictions that successfully correct a large number of pixels. However, this precision may cause a small subset of ambiguous predictions to shift from a `coarse but safe' estimate to a more specialized prediction that falls just outside the wide acceptance margin of $\delta_3$. Ultimately, we interpret this as a strong positive indicator: it demonstrates that MoA-DepthCLIP is successfully learning to make fine-grained, high-precision predictions, a capability strongly reflected by its superior performance on the more challenging $\delta_1$ and $\delta_2$ metrics.

This systematic analysis validates the contribution of each component, demonstrating that the combination of a strong backbone, a composite loss, specialized adapters, and an optimized binning strategy is crucial for achieving high performance.

\section{Ablation Study}
\label{sec:ablation}

We perform ablation studies to examine the effect of two critical hyperparameters in MoA-DepthCLIP: (1) the number of experts used in the Mixture-of-Adapters (MoA) layers, and (2) the number of depth bins used for discretization. This two-stage procedure provides clear, factorized insights and motivates the final hyperparameter choices used in our main experiments.

\subsection{Effect of Expert Count}
In the first stage, we fix the number of depth bins to a coarse value of 10 and vary the number of experts in each MoA layer. As reported in Table~\ref{tab:ablation_experts}, the performance remains relatively stable across different expert counts, suggesting our framework is robust to this hyperparameter. Although the single-expert baseline ($K=1$) performs competitively in terms of AbsRel error, it lacks the multi-expert capacity required for semantic specialization inherent to MoA designs. Conversely, while increasing to $K=8$ yields the best accuracy ($\delta_1 = 0.665$), it doubles the computational overhead for marginal gains. Therefore, we select $K=4$ as a balanced configuration that provides sufficient capacity for expert specialization without the diminishing returns observed at higher counts.

\begin{table}[h]
\centering
\caption{Ablation on the number of experts, with 10 bins. $K=4$ is chosen as a balanced trade-off.}
\label{tab:ablation_experts}
\small
\renewcommand{\arraystretch}{1.1}
\begin{tabular}{c|ccc}
\hline
\textbf{\#Experts} & \textbf{AbsRel $\downarrow$} & \textbf{RMSE $\downarrow$} & $\boldsymbol{\delta_1 \uparrow}$ \\
\hline
1 & \textbf{0.394} & 0.560 & 0.660 \\
2 & 0.396 & 0.559 & 0.661 \\
4 & 0.396 & 0.561 & 0.661 \\
8 & 0.395 & \textbf{0.557} & \textbf{0.665} \\
16 & 0.395 & 0.558 & 0.661 \\
\hline
\end{tabular}
\end{table}

\subsection{Effect of Depth Bin Count}
After fixing the number of experts to 4, we vary the number of depth bins to explore the impact of discretization granularity. Table~\ref{tab:ablation_bins} shows a clear trend. Performance improves as the number of bins increases up to 128, which achieves the best results across all key metrics. Further increasing the bin count to 180 or 200 leads to a slight degradation in performance. We hypothesize that this is due to data sparsity for each bin, making it harder for the model to learn a stable distribution with limited data. This analysis confirms that \textbf{N=128} is the optimal choice for MoA-DepthCLIP.

\begin{table}[h]
\centering
\caption{Ablation on the number of depth bins, with 4 experts. $N=128$ provides the best performance.}
\label{tab:ablation_bins}
\small
\renewcommand{\arraystretch}{1.1}
\begin{tabular}{c|ccc}
\hline
\textbf{\#Bins} & \textbf{AbsRel $\downarrow$} & \textbf{RMSE $\downarrow$} & $\boldsymbol{\delta_1 \uparrow}$ \\
\hline
40  & 0.327 & 0.522 & 0.737 \\
64  & 0.329 & 0.534 & 0.730 \\
128 & \textbf{0.321} & \textbf{0.521} & \textbf{0.745} \\
180 & 0.323 & 0.532 & 0.735 \\
200 & 0.329 & 0.541 & 0.717 \\
\hline
\end{tabular}
\end{table}

\section{Conclusion}

In this paper, we presented MoA-DepthCLIP, a parameter-efficient framework for adapting pretrained CLIP representations for monocular depth estimation. Our approach avoids expensive full-backbone fine-tuning by combining two effective strategies: the insertion of lightweight MoA modules and selective fine-tuning of the final layers of the ViT backbone. The core of our method is a dual-head prediction architecture that simultaneously performs depth classification and regression, processing visual features that have been fused with a global scene context vector.

Our comprehensive experiments and ablation studies validated our design choices, identifying a configuration with 4 experts and 128 depth bins as optimal. A key element of our framework is the composite loss function, which enables the training of this dual-head system by simultaneously supervising a classification head with a cross-entropy loss and a regression head with L1 and SILog losses.

The results confirm the effectiveness of our approach. MoA-DepthCLIP significantly outperforms the DepthCLIP baseline and achieves competitive performance against much larger foundation models, despite using only a fraction of their trainable parameters. This work demonstrates the considerable potential of targeted, lightweight adaptation strategies for bridging the gap between the semantic richness of VLMs and the geometric precision required for dense prediction tasks. Future avenues include extending our framework to diverse outdoor datasets. Furthermore, incorporating dynamic components, like attention-based prompt selection, could further improve the model's performance.

%
%
%
\bibliography{custom}

\begin{thebibliography}{10}
\providecommand{\url}[1]{\texttt{#1}}
\providecommand{\urlprefix}{URL }
\providecommand{\doi}[1]{https://doi.org/#1}

\bibitem{auty2023learning}
Auty, D., Mikolajczyk, K.: Learning to prompt clip for monocular depth estimation: Exploring the limits of human language. In: Proceedings of the IEEE/CVF International Conference on Computer Vision (ICCV) Workshops. pp. 2039--2047 (2023)

\bibitem{bhat2023zoedepth}
Bhat, S.F., Birkl, R., Wofk, D., Wonka, P., M{\"u}ller, M.: Zoedepth: Zero-shot transfer by combining relative and metric depth. In: Proceedings of the IEEE/CVF Conference on Computer Vision and Pattern Recognition (CVPR) (2023)

\bibitem{bhat2021adabins}
Bhat, S.F., Wofk, D., Birkl, R., M{\"u}ller, M., Wonka, P.: {AdaBins}: Adaptive discretization for monocular depth estimation. In: Proceedings of the IEEE/CVF Conference on Computer Vision and Pattern Recognition (CVPR). pp. 14494--14503 (2021)

\bibitem{dosovitskiy2020vit}
Dosovitskiy, A., Beyer, L., Kolesnikov, A., Weissenborn, D., Zhai, X., Unterthiner, T., Dehghani, M., Minderer, M., Heigold, G., Gelly, S., Uszkoreit, J., Houlsby, N.: An image is worth 16x16 words: Transformers for image recognition at scale. In: International Conference on Learning Representations (ICLR) (2021)

\bibitem{eigen2014depth}
Eigen, D., Puhrsch, C., Fergus, R.: Depth map prediction from a single image using a multi-scale deep network. In: Advances in Neural Information Processing Systems (2014)

\bibitem{fu2018deep}
Fu, H., Gong, M., Wang, C., Batmanghelich, K., Tao, D.: Deep ordinal regression network for monocular depth estimation. In: Proceedings of the IEEE conference on computer vision and pattern recognition (CVPR). pp. 2002--2011 (2018)

\bibitem{gao2021clip}
Gao, P., Geng, S., Zhang, R., Ma, T., Fang, R., Zhang, Y., Li, H., Qiao, Y.: Clip-adapter: Better vision-language models with feature adapters. International Journal of Computer Vision (IJCV)  \textbf{132},  581--595 (2024)

\bibitem{houlsby2019adapter}
Houlsby, N., Giurgiu, A., Jastrzebski, S., Morrone, B., De~Laroussilhe, Q., Gesmundo, A., Attariyan, M., Gelly, S.: Parameter-efficient transfer learning for {NLP}. In: International Conference on Machine Learning (ICML). pp. 2790--2799. PMLR (2019)

\bibitem{hu2024learning}
Hu, X., Zhang, C., Zhang, Y., Hai, B., Yu, K., He, Z.: Learning to adapt clip for few-shot monocular depth estimation. In: Proceedings of the IEEE/CVF Winter Conference on Applications of Computer Vision (WACV). pp. 5594--5603 (2024)

\bibitem{kim2024clip}
Kim, S., Kang, J., Kim, D., Lee, S.: Clip can understand depth. arXiv preprint arXiv:2402.03251  (2024)

\bibitem{li2022depthformer}
Li, Z., Chen, Z., Liu, X., Jiang, J.: Depthformer: Exploiting long-range correlation and local information for accurate monocular depth estimation. arXiv preprint arXiv:2203.14211  (2022)

\bibitem{li2023metric3d}
Li, Z., Liu, X., Jiang, J.: Metric3d: Towards zero-shot metric depth estimation. In: Proceedings of the IEEE/CVF International Conference on Computer Vision (ICCV) (2023)

\bibitem{liu2025dm-adapter}
Liu, Y., Liu, Z., Lan, X., Yang, W., Li, Y., Liao, Q.: Dm-adapter: Domain-aware mixture-of-adapters for text-based person retrieval. In: Proceedings of the AAAI Conference on Artificial Intelligence (2025)

\bibitem{mahjourian2018unsupervised}
Mahjourian, R., Wicke, M., Angelova, A.: Unsupervised learning of depth and ego-motion from monocular video using 3d geometric constraints. In: Proceedings of the IEEE Conference on Computer Vision and Pattern Recognition (CVPR) (2018)

\bibitem{baek2024unidepth}
Piccinelli, L., Yang, Y.H., Sakaridis, C., Segu, M., Li, S., Van~Gool, L., Yu, F.: Unidepth: Universal monocular metric depth estimation. In: Proceedings of the IEEE/CVF Conference on Computer Vision and Pattern Recognition (CVPR) (2024)

\bibitem{radford2021learning}
Radford, A., Kim, J.W., Hallacy, C., Ramesh, A., Goh, G., Agarwal, S., Sastry, G., Askell, A., Mishkin, P., Clark, J., Krueger, G., Sutskever, I.: Learning transferable visual models from natural language supervision. In: Proceedings of the International Conference on Machine Learning (ICML). PMLR (2021)

\bibitem{rao2021denseclip}
Rao, Y., Zhao, W., Chen, G., Tang, Y., Zhu, Z., Huang, G., Zhou, J., Lu, J.: Denseclip: Extract free dense labels from clip. arXiv preprint arXiv:2112.01071  (2021)

\bibitem{silberman2012indoor}
Silberman, N., Hoiem, D., Kohli, P., Fergus, R.: Indoor segmentation and support inference from rgbd images. In: European Conference on Computer Vision (ECCV). Springer (2012)

\bibitem{son2024cabins}
Son, E., Lee, S.J.: Cabins: Clip-based adaptive bins for monocular depth estimation. In: Proceedings of the IEEE/CVF Conference on Computer Vision and Pattern Recognition (CVPR) Workshops. pp. 4557--4567 (2024)

\bibitem{wang2022adamix}
Wang, Y., Agarwal, S., Mukherjee, S., Liu, X., Gao, J., Awadallah, A.H., Gao, J.: Adamix: Mixture-of-adaptations for parameter-efficient model tuning. In: Proceedings of the 2022 Conference on Empirical Methods in Natural Language Processing (EMNLP). pp. 5295--5309. Association for Computational Linguistics (2022)

\bibitem{yang2024depthanything}
Yang, L., Kang, B., Huang, Z., Xu, X., Feng, J., Zhao, H.: Depth anything: Unleashing the power of large-scale unlabeled data. In: Proceedings of the IEEE/CVF Conference on Computer Vision and Pattern Recognition (CVPR) (2024)

\bibitem{zeng2018pc}
Zeng, W., Karaoglu, S., Gevers, T.: Inferring point clouds from single monocular images by depth intermediation. In: Proceedings of the European Conference on Computer Vision (ECCV) (2018)

\bibitem{zhang2020unsupervised}
Zhang, M., Ye, X., Fan, X., Zhong, W.: Unsupervised depth estimation from monocular videos with hybrid geometric-refined loss and contextual attention. Neurocomputing  \textbf{379} (2020)

\bibitem{zhang2022pointclip}
Zhang, R., Guo, Z., Zhang, W., Li, K., Miao, X., Cui, B., Qiao, Y., Gao, P., Li, H.: Pointclip: Point cloud understanding by clip. In: Proceedings of the IEEE/CVF Conference on Computer Vision and Pattern Recognition (CVPR) (2022)

\bibitem{zhang2022monodetr}
Zhang, R., Qiu, H., Wang, T., Guo, Z., Tang, Y., Xu, X., Cui, Z., Qiao, Y., Li, H., Gao, P.: Monodetr: Depth-guided transformer for monocular 3d object detection. In: Proceedings of the IEEE/CVF International Conference on Computer Vision (ICCV) (2023)

\bibitem{zhang2022depthclip}
Zhang, R., Zeng, Z., Guo, Z., Li, Y.: Can language understand depth? In: Proceedings of the 30th ACM International Conference on Multimedia. Association for Computing Machinery (2022)

\bibitem{zhang2021tip}
Zhang, R., Zhang, W., Fang, R., Gao, P., Li, K., Dai, J., Qiao, Y., Li, H.: Tip-adapter: Training-free adaption of clip for few-shot classification. In: Proceedings of the European Conference on Computer Vision (ECCV) (2022)

\bibitem{zhou2021learning}
Zhou, K., Yang, J., Loy, C.C., Liu, Z.: Learning to prompt for vision-language models. arXiv preprint arXiv:2109.01134  (2021)

\bibitem{zhou2022detic}
Zhou, X., Girdhar, R., Joulin, A., Krahenbuhl, P., Misra, I.: Detecting twenty-thousand classes using image-level supervision. In: Proceedings of the European Conference on Computer Vision (ECCV). pp. 350--368 (2022)

\end{thebibliography}
%
%
%
\end{document}